%% file: ICDAR26.tex
\documentclass[runningheads]{llncs}
\usepackage[T1]{fontenc}
\usepackage{graphicx}
\usepackage{orcidlink}
\usepackage{listings}
\usepackage{pgfplots}
\usepackage{siunitx}
\usepackage{booktabs}
\usepackage{hyperref}
\usepackage{glossaries}
\pgfplotsset{compat=1.18}
\usepackage{subcaption}

\newacronym{wi}{WI}{Writer Identification}
\newacronym{wr}{WR}{Writer Retrieval}
\newacronym{wv}{WV}{Writer Verification}
\newacronym{cnn}{CNN}{Convolutional Neural Network}
\newacronym{gnn}{GNN}{Graph Neural Network}
\newacronym{lbp}{LBP}{Local Binary Pattern}
\newacronym{lpq}{LPQ}{Local Phase Quantization}
\newacronym{svm}{SVM}{Support Vector Machine}
\newacronym{gscm}{GSCM}{Gray-Scale Co-occurrence Matrices}
\newacronym{wed}{WED}{Weighted Euclidean Distance}
\newacronym{knn}{KNN}{K-Nearest Neighbors}
\newacronym{cc}{CC}{Connected-Component}
\newacronym{sds}{SDS}{SIFT Descriptor Signature}
\newacronym{soh}{SOH}{Scale and Orientation Histogram}
\newacronym{map}{mAP}{Mean Average Precision}
\newacronym{eer}{EER}{Equal Error Rate}
\newacronym{aer}{AER}{Average Error Rate}
\newacronym{gcn}{GCN}{Graph Convolutional Network}
\newacronym{sa}{SA}{Sentiment Analysis}
\newacronym{lstm}{LSTM}{Long Short-Term Memory}
\newacronym{gat}{GAT}{Graph Attention Network}
\begin{document}
\glsdisablehyper
\title{Graph-based Aggregation for Writer Identification and Retrieval}
\author{Oliver Pilizar\inst{1} \and
Marco Peer\orcidlink{0000-0001-6843-0830}\inst{2} \and
Florian Kleber\orcidlink{0000-0001-8351-5066}\inst{1}
}

\authorrunning{O. Pilizar et al.}

\institute{
Computer Vision Lab, TU Wien, Austria  \and
AIBEX Group, University of Fribourg, Fribourg, Switzerland
}
\maketitle              % typeset the header of the contribution
\begin{abstract}
Writer identification and writer retrieval aim to assign handwritten documents to their correct authors or to retrieve documents written by the same individual. While current text-independent, offline approaches achieve high performance at the page level, their accuracy degrades substantially for line- and word-level tasks, partly due to simplistic aggregation methods that fail to preserve fine-grained stylistic information. This paper proposes a graph neural network–based aggregation framework for writer identification and retrieval, representing documents or lines as graphs of local handwriting features with explicitly modeled spatial relationships. A hybrid feature extraction strategy is employed, combining different local descriptors within a unified graph representation. The approach is evaluated with respect to identification accuracy, hard Top-k, and mean average precision.%, and systematically analyzes the impact of feature combinations, graph construction strategies, and graph neural network architectures. The results provide insight into the suitability of graph-based aggregation for capturing subtle handwriting characteristics, particularly in scenarios with limited handwriting content.

\keywords{Writer Retrieval  \and Writer Identification \and Graph Neural Networks.}
\end{abstract}
\input{sections/01_intro}
\input{sections/02_related_work}
\input{sections/03_methodology}

\input{sections/04_results}
\input{sections/conclusion}
\bibliographystyle{splncs04}
\bibliography{references}




% LNCS Homepage, \url{http://www.springer.com/lncs}, last accessed 2023/10/25
% \end{thebibliography}
\end{document}

%% file: sections/01_intro.tex
\section{Introduction}
\label{sec:intro}

\gls{wi} is the task of assigning a given handwritten document to its correct author by comparing it with documents of known writers, whereas \gls{wr} aims to retrieve all documents written by the same author as a provided reference document \cite{koepf2022writer}.

\gls{wi} and \gls{wr} have applications both in historical and modern contexts.
In the historical domain, they enable the identification of anonymous and potentially undated writings, such as papyri and medieval texts \cite{cilia2021papyrow,peer2023feature}. %They also support the reconstruction of travel paths individual writers took by considering the geographical locations in which the documents have been found. Moreover, by identifying the texts authors wrote or copied, insight into their and their patrons' philosophy can be gained \cite{christlein2019icdar}.
In the modern context, \gls{wi} can be applied to forensic scenarios, such as linking documents to potential suspects or to clarify the authorship of disputed writings, while \gls{wr} allows the retrieval of documents from an archive, for instance, within law enforcement agencies \cite{koepf2022writer}.

While \gls{wi}/\gls{wr} are active research fields \cite{peer2025towards,mattick2024evaluating,raven}, most approaches rely on basic aggregation methods, such as sum or mean pooling, to arrive at a representation for entire documents \cite{Peer2023-kv,peer2025towards}. Current \gls{wr} approaches achieve near-perfect \gls{map} on page-level tasks (CVL: 96–97\%, IAM: 92–94\%), but performance drops when less information is available, as in line-level (CVL: 47–69\%, IAM: 56–75\%) and word-level tasks (CVL: 3–9\%, IAM: 5–14\%) \cite{peer2025towards}.
This performance gap may be due to basic aggregation functions discarding fine-grained details and therefore failing to capture subtle handwriting style information, which may prove crucial \cite{he2021gr}, particularly when only small amounts of handwriting are available, such as on the word- and line-level.
Learned aggregation methods using transformers have, however, not been able to surpass current state-of-the-art methods using mean average pooling \cite{mattick2024evaluating}.
To address this limitation, we employ a \gls{gnn} as an aggregation method, which is capable of capturing spatial feature relations. 

% This motivates the following research questions:
% \begin{enumerate}
%     \item How do \glspl{gnn} used as an aggregation method compare to state-of-the-art \gls{cnn}-based approaches for \gls{wi} and \gls{wr} in terms of hard Top-k, \gls{map}, and accuracy?
%     \item Among keypoint coordinates, SIFT descriptors, image patch embeddings, and word embeddings, which feature combination provides the most accurate identification and retrieval results based on hard Top-k, \gls{map}, and accuracy?
%     \item How do different graph-building strategies influence the hard Top-k, \gls{map}, and accuracy metrics?
%     \item How do different \gls{gnn} architectures affect the hard Top-k, \gls{map} and accuracy metrics?
% \end{enumerate}

The approach presented in this work represents handwritten documents, or individual lines of documents, as graphs, where each node corresponds to a combination of features extracted from a specific keypoint, and edges capture the spatial relationships between these nodes. This approach is partially inspired by sentiment analysis, in which local tokens are aggregated to predict a global meaning, a task to which \glspl{gnn}, among other neural network architectures, have successfully been applied \cite{zhang2019aspect,liao2021multi}. Similarly, in \gls{wi}/\gls{wr} local handwriting features are aggregated to represent an entire line or document, enabling both the identification of the writer and the retrieval of documents written by the same author. 
\glspl{gnn} are particularly suited for this task, as they can explicitly model spatial relationships between individual features by leveraging the underlying graph structure. %This allows them to capture influences such as neighboring letters and strokes. Furthermore, representing nodes as combined feature vectors provides flexibility to test different feature combinations without altering the overall pipeline and graph structure. 
Lastly, the inherent ability of \glspl{gnn} to handle variable input sizes makes the method applicable to both document-level and line-level representations, and allows for the use of even smaller document subsets when needed.

Thus, the following contributions are made:
\begin{itemize}
    \item Application of \glspl{gnn} as an aggregation strategy for offline, text-independent \gls{wi}/\gls{wr}, with a focus on line-level tasks
    \item Evaluation of different feature combinations to determine their impact on identification and retrieval performance
    \item Assessment of different graph-building strategies and comparison of different \gls{gnn} architectures %to highlight the effect of connection methods on identification and retrieval performance
\end{itemize}

The rest of this paper is structured as follows. Section 2 presents related work, %organized into handcrafted and deep learning features, \gls{gnn} approaches in related handwriting domains as well as in sentiment analysis, and learned aggregation functions. 
Section 3 describes the proposed methodology, %, including pre-processing, graph construction, the design of the \gls{gnn} and its different layer types, as well as an overview of the datasets. 
and Section 4 presents the experimental results.%, including comparisons between features, graph building methods, and network architectures. Finally, Section 5 concludes the thesis.%, summarizing the main findings and discussing the potential for future work.

%% file: sections/02_related_work.tex
\section{Related Work}
\label{sec:related_work}
% \gls{wi} and \gls{wr} play an important role in both the historical and modern domains. In the historical context, it enables the attribution of texts to known writers or the reconstruction of travel paths individual writers took by considering the geographical locations in which the documents have been found. The handling of factors such as degradation or ornaments presents additional challenges. In the modern context, forensic analysis allows, for instance, the linking of documents to possible suspects. 
Both \gls{wi} and \gls{wr} encompass a variety of techniques across different aspects of the process. Kaur and Kumar \cite{Kaur2023-cu} define these aspects as feature extraction, feature selection, classification, and distance measures. Kumar et al. \cite{kumar2022offline} state that the feature extraction process of \gls{wi} can be divided into three categories, namely texture, shape, and deep-learning approaches. Additionally, combined approaches leverage both learned and handcrafted features. %This section introduces these categories, including combined methods, with examples from prior works. Afterwards, t

\paragraph{Graph Neural Networks Approaches in Related Handwriting Domains}
\glspl{gnn} have largely remained unexplored in \gls{wi} adjacent domains. An exception is the work by Roy et al. \cite{roy2023offline}, who propose an offline signature verification system employing \glspl{gnn} at the node level.

In their approach, the graph is constructed from the features representing the signature samples. A GraphSAGE architecture is used to aggregate neighboring node information to classify signatures as genuine or forged. The method is evaluated on the MCYT-75 \cite{ortega2003mcyt} and UTSig \cite{soleimani2017utsig} datasets.

Although the \gls{gnn} is applied at the node level rather than to whole-graph representations, this work demonstrates the potential of relational modeling in handwriting analysis and suggests that \glspl{gnn} can be adapted for \gls{wi} and \gls{wr}. 

Recently, the work of Priya et al. \cite{priya} proposes a new dataset for WI, and their method relies on Delauney triangulation, whose graph is forward to a graph convolutional network to classify the writer of an word image.

\paragraph{Learned Aggregation Functions}

Contemporary approaches to feature aggregation into global document representations in the \gls{wi} and \gls{wr} domains typically rely on simplistic aggregation methods such as mean or sum pooling. Mattick et al. \cite{mattick2024evaluating} explore the potential of learned feature aggregators.
Their pipeline begins with the extraction of 32x32 image patches around SIFT keypoints. A standard ResNet-18 is then used to compute their embeddings, which are aggregated into a global image representation via one of two architectures - NetVLAD \cite{arandjelovic2016netvlad}, or a transformer \cite{vaswani2017attention} architecture, which applies self-attention to weight relationships between all patch embeddings, before creating a learned global representation.

% \begin{itemize}
%     \item NetVLAD \cite{arandjelovic2016netvlad}, which aggregates local embeddings by computing the residuals to a set of learned cluster centers.
%     \item A transformer \cite{vaswani2017attention} architecture, which applies self-attention to weight relationships between all patch embeddings, before creating a learned global representation.
% \end{itemize}
% To train the model, several learning objectives are compared:
% \begin{itemize}
%     \item Supervised classification, using cross-entropy loss with center loss regularization to predict writer IDs.
%     \item Unsupervised VICReg \cite{bardes2021vicreg}, which aims to maximize the latent space’s representational capacity, ensuring each embedding dimension carries unique, non-redundant information.
%     \item InfoNCE \cite{musgrave2020metric}, which is a metric learning strategy that optimizes the retrieval likelihood for nearest-neighbor clusters, encouraging discriminative writer-specific features, and mirrors the discrete top-writer retrieval problem
%     \item Triplet loss, which maximizes the distance between negative pairs while minimizing the distance between positive pairs in the embedding space.
% \end{itemize}
Experiments are conducted on the Historical-WI dataset \cite{fiel2017icdar2017}. Performance is evaluated using \gls{map} and precision.
Although the learned aggregation methods aim to capture more complex relationships between patch embeddings, the authors find that they fall short in comparison to state-of-the-art approaches using mean pooling.

These results highlight the importance of explicitly modeling relationships between local features, a capability that \glspl{gnn} naturally provide.
%In this thesis, several types of convolutional \gls{gnn} layers are evaluated to propagate relational feature information, as described in Section \ref{sec:layer_types}.

%% file: sections/03_methodology.tex
\section{Methodology}
\label{sec:methodology}
This section presents the complete pipeline from data input to the final writer identification and retrieval results. %A high-level overview is given in Figure \ref{fig:methodology_pipeline}. 
% Firstly, the datasets used are described. Subsequently, the pre-processing step is outlined, which includes the extraction of text patches based on SIFT keypoints and the retrieval of ground truth information, followed by the creation of the graphs from this data. Lastly, the employed \gls{gnn} architecture, including an explanation of each convolutional layer type used, is given, alongside a description of the writer retrieval and identification step. These descriptions illustrate the individual stages in the pipeline, intermediate results, and statistics about the data.

% \begin{figure}
%   \centering
%   \includegraphics[width=\columnwidth]{graphics/pre-processing/diagrams/GNNs_For_Writer_Retrieval_Id-methodology.drawio.png}
%   \caption{Base overview of the complete pipeline from data input to writer identification/retrieval results}
%   \label{fig:methodology_pipeline}
% \end{figure}

\subsection{Datasets}
As dataset, the CVL \cite{CVL} dataset consisting of 310 different writers has been used. In the default CVL train/test split 27 authors are part of the training set and the other 283 are part of the test set. Since during the experiments, writer retrieval as well as writer identification is tested, the CVL train set is further split into train/validation/test (70\%/15\%/15\%) sets, by randomly splitting each writer's documents according to those percentages, thereby keeping all 27 authors in each set.
This additional split of the training data is necessary to enable the model to be used for both \gls{wi} and \gls{wr}. This newly created test set from the original training data is used exclusively to evaluate identification, while the validation set is used for early stopping. The CVL test set is kept unchanged, as it is already prepared for retrieval and not required for identification.
The inputs used for the experiments are the individual word images as well as their ground truths.

The second dataset used is the IAM \cite{IAM} dataset, consisting of 657 writers, all contributing a varying number of documents. It is not directly used in the \gls{gnn} setups, but rather as an additional option for the training of the \gls{cnn}, which provides the image patch embeddings for the \gls{gnn}.
% To prepare for the training of the \gls{cnn} and to remove the high imbalance of texts available between authors, only a maximum of 100 words per author is chosen, as that is close to the median number of words per author. The result of this balancing step is illustrated in Figure \ref{fig:iam_cap}.

% \begin{figure}
%   \centering
%   \includegraphics[width=\columnwidth]{graphics/pre-processing/images_per_author_after_cap.png}
%   \caption{Illustration of the number of images per author after being capped to a maximum of 100, as well as the mean, median, minimum, and maximum.}
%   \label{fig:iam_cap}
% \end{figure}

\subsection{Pre-Processing and Feature Extraction}
\begin{figure}[t]
  \centering

  \begin{subfigure}{0.48\columnwidth}
    \centering
    \includegraphics[width=\linewidth]{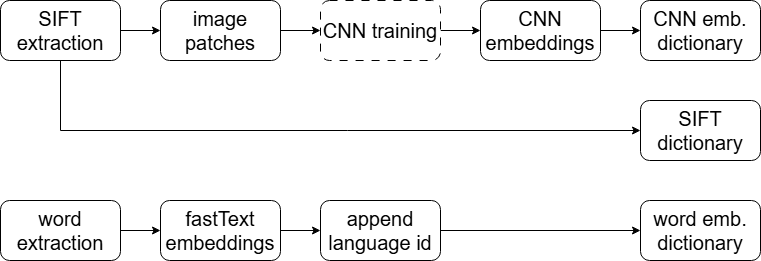}
    \caption{Illustrated pipeline of the pre-processing phase.}
    \label{fig:pre-processing-pipeline}
  \end{subfigure}
  \hfill
  \begin{subfigure}{0.48\columnwidth}
    \centering
    \includegraphics[width=0.8\linewidth]{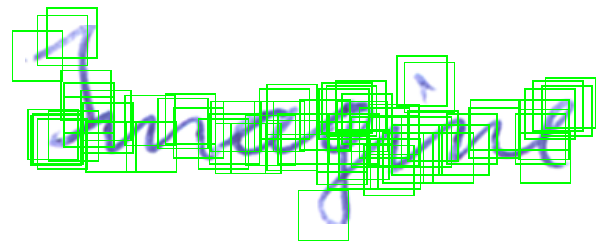}
    \caption{Contours of image patches extracted based on SIFT keypoints, overlayed on the original image to highlight coverage.}
    \label{fig:patch_coverage}
  \end{subfigure}

  \caption{Pre-processing pipeline and patch coverage visualization.}
  \label{fig:preprocessing_combined}
\end{figure}

The goal of this phase is to obtain local features that are later used for \gls{wr} and \gls{wi} within the \gls{gnn} setup. These include \gls{cnn} embeddings of image patches, SIFT features, and textual data in the form of word embeddings. The usage of image patch embeddings in connection with SIFT features is inspired by the work of Christlein et al. \cite{Christlein-unsup}. %, which is introduced in Section \ref{sec:deep_learning}. 
The pre-processing pipeline, which is shown in Figure \ref{fig:pre-processing-pipeline}, consists of two parallel branches to extract the aforementioned data.

The upper branch begins with the extraction of SIFT features, as they are located directly at the strokes, which subsequently serve both as part of the \gls{gnn}'s input feature vector, but also as center points for the creation of the 32x32 pixel image patches in the following step. Figure \ref{fig:patch_coverage} depicts coverage of the word images. These created image patches are then used to train a \gls{cnn}, specifically a ResNet-20, to discriminate between the authors associated with the texts. The objective is to generate embeddings that effectively represent the visual characteristics of the handwriting. Once trained, the \gls{cnn} is used to extract embeddings from the patches, which are saved as key-value pairs to allow efficient access during the graph-building process as part of the \gls{gnn} setup. Simultaneously, the SIFT features are saved in the same format.

In parallel, the lower branch of the pre-processing pipeline extracts the ground truths of all words occurring in the documents. Each word is then embedded using pre-trained fastText vectors, following the approach by Bojanowski et al. \cite{fasttext}. To differentiate between English and German words, an additional identifier is appended to each word embedding. By providing the GNN the ground truths, the model can learn to differentiate how different authors write the same words. Additionally, by adding the language identifier, the model is able to capture how the writing style of an author may vary across languages.

% \begin{figure}
%   \centering
%   \includegraphics[width=0.7\columnwidth]{graphics/pre-processing/diagrams/GNNs_For_Writer_Retrieval_Id-pre_processing.drawio.png}
%   \caption{Illustrated pipeline of the pre-processing phase.}
%   \label{fig:pre-processing-pipeline}
% \end{figure}

% \begin{figure}
%   \centering
%     {\includegraphics[width=0.7\linewidth]{graphics/pre-processing/Imagine_SIFT_Patches.pdf}}
%     \caption{Contours of image patches extracted based on SIFT keypoints, overlayed on the original image to highlight coverage.}
%     \label{fig:patch_coverage}
% \end{figure}

\subsubsection{Convolutional Neural Network using Image Patches}
\label{subsec:CNN_Img_Patches}
To extract the embeddings of the image patches, a ResNet-20 architecture is employed. This selection is based on its input size compatibility with the 32x32 image patches and its output embedding dimension of 64. This dimension provides a balanced contribution to the overall feature vector, which may also incorporate SIFT descriptors (128 dimensions), SIFT keypoint coordinates (2 dimensions), and a word embedding (101 dimensions).

\subsection{Graph Building}
\label{sec:graph_building}
\begin{figure}[t]
    \centering

    % First row: Full and KNN side-by-side
    \subfloat[]{%
        \includegraphics[width=0.45\columnwidth]{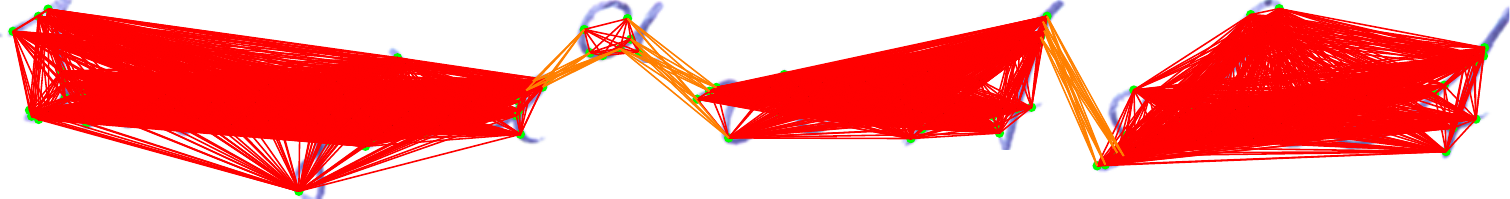}%
        \label{fig:full_vis}%
    }\hfill % Adds horizontal space to push them to the edges
    \subfloat[]{%
        \includegraphics[width=0.45\columnwidth]{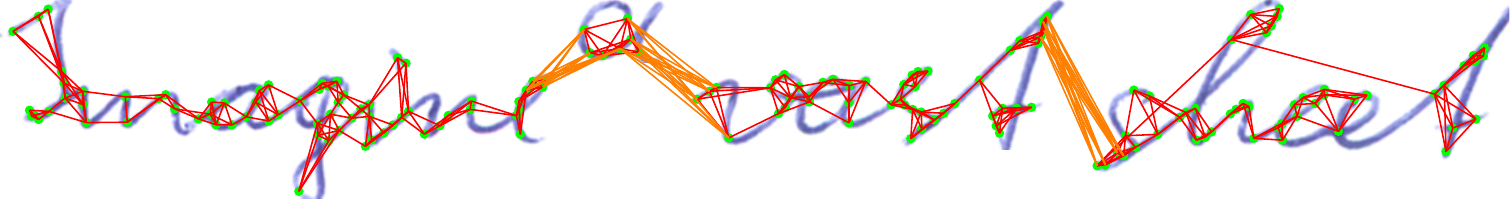}%
        \label{fig:knn_vis}%
    }\\[2mm] % Line break to start the next row

    % Second row: 1Line centered
    \subfloat[]{%
        \includegraphics[width=0.45\columnwidth]{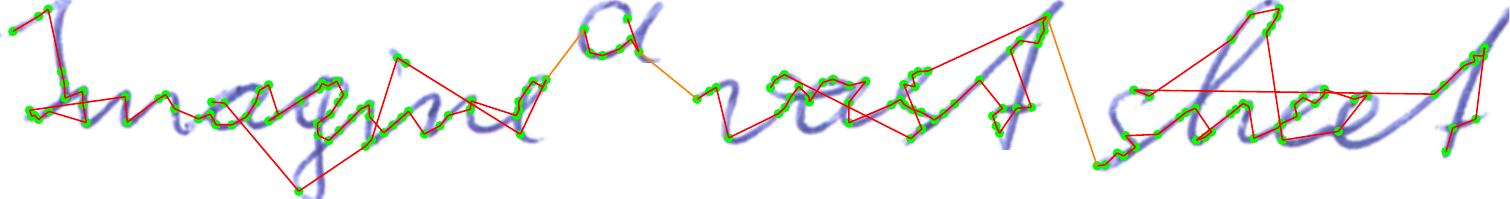}%
        \label{fig:1line_vis}%
    }

    \caption{Line-level depiction of the three connection modes: (a) Full, (b) KNN, (c) 1Line. Orange edges indicate connections between different words.}
    \label{fig:connection_modes_graphs}
\end{figure}

During the graph-building phase, graphs are constructed to represent either individual lines or entire documents. The prepared data, consisting of SIFT feature information, image patch embeddings, and word embeddings, is used to create nodes. To differentiate between the representation of individual lines or entire documents, the data retrieved in this step corresponds to either the currently processed line or the document as a whole, respectively. The process is otherwise identical. Each node is a concatenated feature vector of the aforementioned data, which is always done in the same order. These nodes are then connected via one of three methods: Full, KNN or 1Line. Graphs illustrating these connection method are shown in Figure \ref{fig:connection_modes_graphs}.

\subsubsection{Full Connection Method}
Let $W = \{w_1, w_2, \dots, w_N\}$ be the sequence of $N$ words., and for each word \(w_i\), let \(V_w\) be its set of nodes. Let \(E\) be the set of edges of the graph.

Fully connect nodes within each word:
\begin{equation}
E = \{\, (u,v) \mid w \in W,\; u, v \in V_w,\; u \neq v \,\}
\end{equation}

Connect consecutive words via their outer nodes:
\begin{equation}
\begin{split}
E = \bigl\{\, (u, v) \;\mid 
i \in \{1, \dots, N-1\},\\
u \in \text{rightmost}(V_{w_i}), v \in \text{leftmost}(V_{w_{i+1}})
\,\bigr\}
\end{split}
\end{equation}

The connection mode "Full" creates fully connected subgraphs that represent words, while sequentially connecting them. %Graphs illustrating the "Full" connection method for a line and a single word are shown in Figure \ref{fig:full_graphs}.

\subsubsection{KNN Connection Method}
Let $W = \{w_1, w_2, \dots, w_N\}$ be the sequence of $N$ words, and for each word \(w_i\), let \(V_w\) be its set of nodes. Let \(E\) be the set of edges of the graph. Let \(\text{KNN}_k(u)\) be the set of \(k\)-nearest neighbors of node \(u \in V_w\) within the same word, \(\text{Components}(V_w, E_w)\) the connected components of the word subgraph, and \(\text{repr}(C)\) a representative node from component \(C\), chosen arbitrarily, as any node suffices to connect the component.

Connect each node to its \(k\)-nearest neighbors within the word:
\begin{equation}
E = \{\, (u,v) \mid w \in W,\; u \in V_w,\; v \in \text{KNN}_k(u) \,\}
\end{equation}

Ensure each word subgraph is connected by linking its components sequentially:
\begin{equation}
\begin{split}
E = \bigl\{\, 
(\text{repr}(C_i), \text{repr}(C_{i+1})) \;\mid 
w \in W,\\
C_i, C_{i+1} \in \text{Components}(V_w, E_w)
\,\bigr\}
\end{split}
\end{equation}

Connect consecutive words via their outer nodes:
\begin{equation}
\begin{split}
E = \{\, (u, v) \mid & \; i \in \{1, \dots, N-1\},\\
                       & \; u \in \text{rightmost}(V_{w_i}),\\
                       & \; v \in \text{leftmost}(V_{w_{i+1}}) \,\}
\end{split}
\end{equation}

The "KNN" connection method aims to reduce the number of edges by connecting each node only to its close neighbors, utilizing the k-nearest-neighbor algorithm. The value $k=4$ is chosen to allow the formation of natural clusters within words while preventing the creation of fully connected graphs.

\subsubsection{1Line Connection Method}
Let $W = \{w_1, w_2, \dots, w_N\}$ be the sequence of $N$ words, \(V_w = \{v_1, v_2, \dots, v_{n_w}\}\) the ordered set of nodes for word \(w_i\), and \(E\) the set of edges. Let \(\text{dist}(u,v)\) denote the distance between two nodes, and \(\text{first}(V_w)\) and \(\text{last}(V_w)\) the first and last nodes of word \(w\).

Connect nodes sequentially within a word by always linking the closest unconnected node:

Let \(C_w \subseteq V_w\) be the set of connected nodes, initialized with the top-left node of the first word in the sequence \(C_w = \{\text{first}(V_{w_1})\}\). Then, iteratively, connect each remaining node in order of proximity to the last added node \(u_{\text{last}} \in C_w\):
\begin{equation}
\begin{split}
E = \bigl\{\, (u_{\text{last}}, v) \;\mid & \;
w \in W, v \in V_w \setminus C_w,\\
& v = \arg\min_{x \in V_w \setminus C_w} \text{dist}(x, u_{\text{last}})
\,\bigr\}
\end{split}
\end{equation}

Repeat until all nodes in \(V_w\) are in \(C_w\).

Connect consecutive words via the first and last nodes:
% \[
% \forall i \in \{2, \dots, N\}: \quad (\text{last}(V_{w_{i-1}}), \text{first}(V_{w_i})) \in E
% \]
\begin{equation}
E = 
\left\{
\left( \text{last}(V_{w_{i-1}}),\ \text{first}(V_{w_i}) \right)
\;\middle|\;
i \in \{2, \dots, N\}
\right\}
\end{equation}

The "1Line" connection method minimizes the number of edges while maintaining a connected graph. Instead of trying to replicate the original line stroke order, this method aims to be consistent for repeated samples of the same word. %Graphs illustrating the "1Line" connection method for a line and a single word are shown in Figure \ref{fig:1Line_graphs}.

\subsection{Graph Neural Networks}
\begin{figure}[t]
    \centering

    \begin{subfigure}[c]{0.29\columnwidth}
        \centering
        \includegraphics[width=\linewidth]
        {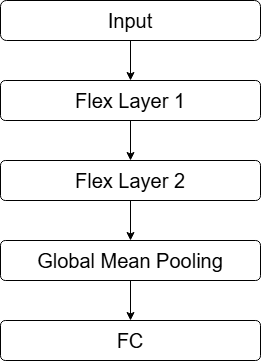}
        \caption{}
        \label{fig:GNN_general}
    \end{subfigure}
    \hfill
    \begin{subfigure}[c]{0.69\columnwidth}
        \centering
        \includegraphics[width=\linewidth]
        {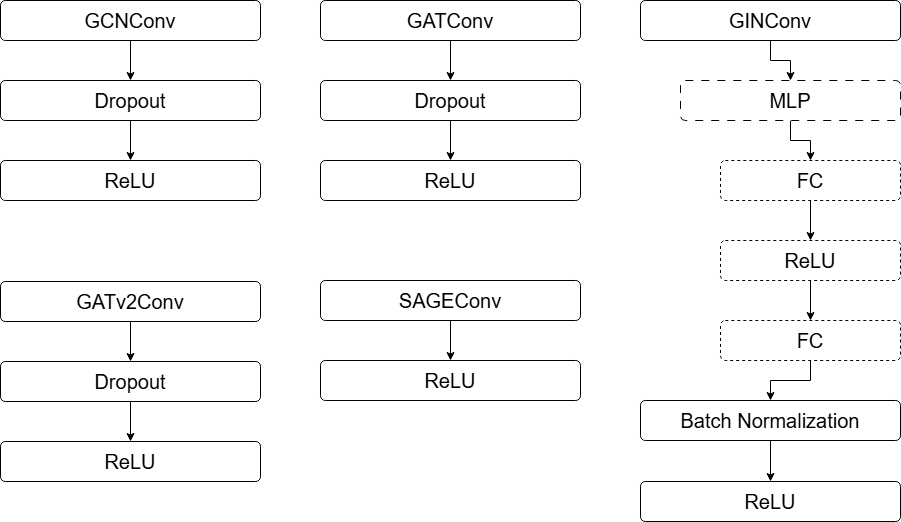}
        \caption{}
        \label{fig:GNN_flex_layers}
    \end{subfigure}

    \caption{Overview of the \gls{gnn} architecture showing the general architecture in (a) and the detailed architecture of the Flex Layers used to test different setups in (b).}
    \label{fig:gnn_architecture}
\end{figure}

The general \gls{gnn} architecture used for this work is shown in Figure \ref{fig:GNN_general}. Figure \ref{fig:GNN_flex_layers} further describes the Flex Layers appearing in the general architecture, which are configurable convolutional layers.
The input for the \gls{gnn} is the graphs created in the previous step, which consist of the nodes representing combined feature vectors and edges denoting connections within and between words in different ways, depending on the chosen connection method as introduced in Section \ref{sec:graph_building}.
Afterwards the Flex Layers are employed, which can be stacked any number of times. The first layer optionally changes the hidden feature dimension with subsequent Flex Layers continuing with the same size. By default, two Flex Layers are used with hidden dimensions equal to the input size. A global mean pooling layer is then employed to aggregate node features into a global representation of the graph.
For \gls{wi} the predicted author corresponds to the class with the highest output probability of the fully connected layer, while for \gls{wr} embeddings from before the fully connected layer are compared using Euclidean distance.

\subsection{Layer Types}
\label{sec:layer_types}

Graph neural network architectures commonly incorporate diverse layer types, such as convolutional, pooling, and aggregation layers. This paper investigates mainly convolutional layers. Specifically, GCNConv \cite{GCNConv}, GATConv \cite{GATConv}, GATv2\-Conv \cite{GATv2Conv}, SAGEConv \cite{SAGEConv} and GINConv \cite{GINConv} have been tested.
% Specifically, the following types are tested:
% \begin{itemize}
%   \item GCNConv \cite{GCNConv}
%   \item GATConv \cite{GATConv}
%   \item GATv2Conv \cite{GATv2Conv}
%   \item SAGEConv \cite{SAGEConv}
%   \item GINConv \cite{GINConv}
% \end{itemize}

GCNConv is used as the standard convolutional layer during the experiments that do not specifically test the performance and viability of the different layer implementations, but, for instance, the importance of different features extracted from the documents.

In the context of writer identification and retrieval GCNConv means finding localized patterns in the provided features that characterize the writer's style, such as recurring traits in the way characters are written via SIFT and the image patch embeddings, as well as the additional contextual information about the word itself via the word embeddings.

GATConv implements an attention mechanism, which allow the network to gain a deeper understanding and highlight the most important feature nodes accordingly. For instance, anomalies in a writer's style may be assigned heavier weights, while common features or ones that do not carry as much information, for instance, the dot of a lower case "i", may get assigned lower weights. This should, in theory, allow for a more refined representation of each author's individual style. GATv2Conv is an improved version of the attention mechanism and it is analyzed if more context-dependent patterns can be found, and the overall performance of the model can thereby be improved. SAGEConv is designed to learn a function that can identify unique patterns for each writer, which can then be transferred to completely new and unseen authors. GINConv has a highly discriminative representation, which is supposed to allow the model to differentiate between authors with very similar writing styles.

%% file: sections/04_results.tex
\section{Results}

% \subsection{General}
%This chapter presents the evaluation protocol and the experimental results of the methodology. 
The primary focus of the evaluation lies on \gls{wr} on a per-line basis, as this remains the more challenging task and provides more room for improvement compared to document-level retrieval, where state-of-the-art methods achieve over 96\% \gls{map} \cite{peer2025towards}.
%Document-level experiments are additionally included to provide a complete comparison to existing methods and to demonstrate the adaptability of the proposed graph-based approach, which can be adapted to either level by varying whether the constructed graph includes all nodes of a document or only those belonging to individual lines.
%\gls{wi} is explored with fewer experiments than \gls{wr}, as the task is comparatively simpler and the demonstrated tests already achieve test accuracies of around 97\% and up.

%The initial experiments investigate different parameter aspects of the proposed \gls{wr} approach to identify the most effective configurations. These include the types of convolutional layers used within the \gls{gnn}, different combinations of extracted features, as well as multiple graph construction strategies.
%Subsequent experiments address comparisons between line-level and document-level retrieval, as well as the challenge of lower sample numbers during document-level tasks. Lastly, variations to the training data are explored to assess their potential impact on performance, concluding the \gls{wr} section with a custom layout that builds on previous findings.

%Subsequent experiments focus on \gls{wi}, following the same general structure, but are limited to convolutional layer types within the \gls{gnn}, feature combinations, and graph connection methods. 

\subsection{Retrieval}
This section investigates the performance of \gls{wr}, using the CVL dataset by default, unless explicitly stated otherwise. The official train/test split is used, with 70\% of the training set reserved for training, 15\% for validation, and 15\% for testing.
%This additional split of the training data is necessary to enable the model to be used for both \gls{wi} and \gls{wr}. This newly created test set from the original training data is used exclusively to evaluate identification, while the evaluation set is used for early stopping in both tasks. The CVL test set is kept unchanged, as it is already prepared for retrieval and not required for identification.
\begin{table}[t]
\centering

\begin{subtable}[b]{0.32\textwidth}
\centering
\begin{tabular}{lr}
\toprule
Learning Rate & 0.001 \\
Batch Size & 32 \\
Epochs & 250 \\
Patience & 15 \\
Weight Decay & 0.0001 \\
Loss & CE \\
Optimizer & Adam \\
\bottomrule
\end{tabular}
\caption{Training}
\end{subtable}
\hfill
\begin{subtable}[b]{0.32\textwidth}
\centering
\begin{tabular}{lr}
\toprule
Hidden dim. & Input dim \\
Num. of Layers & 2 \\
Layer Type & GCN \\
Features & SKPW \\
Connection & 1line \\
Scope & line-level \\
\bottomrule
\end{tabular}
\caption{Model Architecture}

\end{subtable}
\hfill
\begin{subtable}[b]{0.32\textwidth}
\centering
\begin{tabular}{lr}
\toprule
SIFT Descriptor (S) & 128 \\
SIFT Keypoint (K) & 2 \\
Patch Embedding (P) & 64 \\
Word Embedding (W) & 101 \\
\bottomrule
\end{tabular}
\caption{Feature Dimensions}

\end{subtable}

\caption{Default hyperparameters and feature dimensions used to train the \gls{gnn}. Feature dimensions are shown as reference for calculating layer dimensions.}
\label{tab:training_config}
\end{table}

Table \ref{tab:training_config} shows the standard configuration of the experiments. They stay the same through all of them unless otherwise specified, or the purpose of the experiment is to vary a certain parameter to draw conclusions on its impact.
To get the input feature dimension of the \gls{gnn}, we add up all the individual feature vector dimensions used for that \gls{gnn}. The sub-table \textit{Feature Dimensions} gives an overview of the possible features included, their notations (S - SIFT Descriptor, K - SIFT Keypoint, P - Image Patch Embedding, W - Word Embedding), and their dimensions. When combining features such as SKPW, the individual feature vectors are concatenated, resulting in a total dimensionality equal to the sum of their individual dimensions, which in the case of SKPW is 295. The default for the experiments is to use two hidden layers that both have the same dimension as the input layer. %More information about the used features themselves and the other parameters being tested is given in their respective sections.

\subsubsection{Layer Types - GCN, GAT, GATv2, SAGE, GIN}
\label{subsec:layer_types}
Different layer types of the \gls{gnn} are evaluated and in each experiment, the number of layers and the dimensions remain fixed, according to the overview in Table \ref{tab:training_config} . %, only the type of layer itself changes. 
For the GAT and GATv2 layers, the number of multi-head attentions is set to one. Each GIN layer uses an underlying two-layer MLP with a ReLU activation function in between. After each GCN, GAT, and GATv2 layer, a dropout layer with a 50\% probability is added, as in the original papers for the GCN \cite{GCNConv} and GAT \cite{GATConv}, dropout layers are used. In contrast, the original papers apply dropout before the convolutional layers, which in the experiments conducted here resulted in a decrease of up to approximately 10\% \gls{map}. The GATv2 paper \cite{GATv2Conv} notes that dropout does not have a significant influence on the results, which is also the case for the results of the experiments conducted here. The results shown for GATv2 are the ones using dropout layers. SAGE and GIN do not include dropout layers for the experiments, as the results without are better, however, instead of dropout layers, a batch-norm is used after each GINConv layer. Table \ref{tab:retrieval_layer_types} shows those results.

% \begin{table}
%   \centering
%   \scriptsize
%   \begin{tabular}{lc|ccccc}
%     \toprule
%     & \multicolumn{1}{c|}{Training} & \multicolumn{5}{c}{Retrieval (\%)} \\
%     \cmidrule{2-7}
%     & \# Epochs & mAP & Top-1 & Top-3 & Top-5 & Top-10 \\
%     \midrule
%     GCN    & 53  & 53.69 & 86.33 & 69.92 & 58.40 & 38.58 \\
%     GAT    & 53 & 52.56 & 85.59 & 68.50 & 56.42 & 36.97 \\
%     GATv2  & 93 & 55.93 & 87.36 & 71.87 & 60.69 & 41.67 \\
%     SAGE   & 64 & 54.41 & 86.02 & 69.11 & 57.69 & 39.12 \\
%     \textbf{GIN}    & \textbf{82} & \textbf{60.24} & \textbf{90.46} & \textbf{77.65} & \textbf{67.53} & \textbf{49.48} \\
%     \bottomrule
%   \end{tabular}
%   \caption{Performance metrics of different layer types}
%   \label{tab:retrieval_layer_types}
% \end{table}
% \begin{table}
%   \centering
%   \scriptsize
%   \begin{tabular}{l|ccccc}
%     \toprule
%     & \multicolumn{5}{c}{Retrieval (\%)} \\
%     \cmidrule{2-6}
%     & mAP & Top-1 & Top-3 & Top-5 & Top-10 \\
%     \midrule
%     GCN    & 53.69 & 86.33 & 69.92 & 58.40 & 38.58 \\
%     GAT    & 52.56 & 85.59 & 68.50 & 56.42 & 36.97 \\
%     GATv2  & 55.93 & 87.36 & 71.87 & 60.69 & 41.67 \\
%     SAGE   & 54.41 & 86.02 & 69.11 & 57.69 & 39.12 \\
%     \textbf{GIN} & \textbf{60.24} & \textbf{90.46} & \textbf{77.65} & \textbf{67.53} & \textbf{49.48} \\
%     \bottomrule
%   \end{tabular}
%   \caption{Performance metrics of different layer types.}
%   \label{tab:retrieval_layer_types}
% \end{table}

The results show that all layer types, including GCN, which is used as the standard for the experiments, perform on a similar level, except for GIN, which achieves about 7\% higher \gls{map} compared to GCN. Layer types within about a percentage point can be considered effectively equivalent.

\subsubsection{Feature Combinations - S/K/P/W}
\label{subsec:feature-combinations}

Individual and combined features used during training and inference, with the notation of S, K, P, W described in Table \ref{tab:training_config}, sub-table “Feature Dimensions" are evaluated. The results are shown in Table \ref{tab:retrieval_feature_combinations} and are grouped by the number of features used. %The best performing one in each group is highlighted.

\begin{table}[t]
\centering
\scriptsize

\begin{minipage}[b]{0.48\linewidth}
\centering

\begin{subtable}[b]{\linewidth}
\centering
\begin{tabular}{lccccc}
  \toprule
  & \multicolumn{5}{c}{Retrieval (\%)} \\
  \cmidrule{2-6}
  & mAP & Top-1 & Top-3 & Top-5 & Top-10 \\
  \midrule
  GCN    & 53.69 & 86.33 & 69.92 & 58.40 & 38.58 \\
  GAT    & 52.56 & 85.59 & 68.50 & 56.42 & 36.97 \\
  GATv2  & 55.93 & 87.36 & 71.87 & 60.69 & 41.67 \\
  SAGE   & 54.41 & 86.02 & 69.11 & 57.69 & 39.12 \\
  \textbf{GIN} & \textbf{60.24} & \textbf{90.46} & \textbf{77.65} & \textbf{67.53} & \textbf{49.48} \\
  \bottomrule
\end{tabular}
\caption{Layer types}
\label{tab:retrieval_layer_types}
\end{subtable}

\vspace{0.8em}

\begin{subtable}[b]{\linewidth}
\centering
\begin{tabular}{lccccc}
  \toprule
  & \multicolumn{5}{c}{Retrieval (\%)} \\
  \cmidrule{2-6}
  & mAP & Top-1 & Top-3 & Top-5 & Top-10 \\
  \midrule
  \textbf{1Line}  & \textbf{53.77} & \textbf{86.12} & \textbf{70.00} & \textbf{58.34} & \textbf{38.55} \\
  KNN    & 51.00 & 84.33 & 65.62 & 53.35 & 33.30 \\
  Full   & 36.18 & 68.50 & 44.01 & 31.51 & 15.92 \\
  \bottomrule
\end{tabular}
\caption{Connection methods}
\label{tab:retrieval_connection_modes}
\end{subtable}

\end{minipage}
\hfill
\begin{minipage}[b]{0.48\linewidth}
\centering

\begin{subtable}[b]{\linewidth}
\centering
\begin{tabular}{lccccc}
  \toprule
  & \multicolumn{5}{c}{Retrieval (\%)} \\
  \cmidrule{2-6}
  & mAP & Top-1 & Top-3 & Top-5 & Top-10 \\
  \midrule
  S    & 9.74 & 20.43 & 5.02 & 2.33 & 0.69 \\
  K    & 1.20 & 1.31 & 0.00 & 0.00 & 0.00 \\
  \textbf{P}    & \textbf{58.79} & \textbf{89.76} & \textbf{76.06} & \textbf{65.97} & \textbf{47.29} \\
  W    & 0.35 & 0.01 & 0.00 & 0.00 & 0.00 \\
  \midrule
  SK   & 6.28 & 15.75 & 2.87 & 1.15 & 0.20 \\
  \textbf{SP}   & \textbf{60.09} & \textbf{90.55} & \textbf{77.41} & \textbf{67.70} & \textbf{48.85} \\
  SW   & 9.58 & 19.54 & 4.42 & 2.05 & 0.57 \\
  KP   & 53.17 & 85.28 & 68.08 & 56.54 & 37.96 \\
  KW   & 2.02 & 3.55 & 0.05 & 0.01 & 0.00 \\
  PW   & 59.58 & 89.78 & 76.07 & 66.23 & 48.04 \\
  \midrule
  SKP  & 55.31 & 86.78 & 70.28 & 59.22 & 39.70 \\
  SKW  & 8.07  & 19.54 & 4.08  & 1.74  & 0.27 \\
  \textbf{SPW}  & \textbf{57.96} & \textbf{89.74} & \textbf{75.65} & \textbf{64.89} & \textbf{46.21} \\
  KPW  & 52.05 & 85.14 & 67.42 & 55.64 & 36.12 \\
  \midrule
  SKPW & 52.31 & 84.68 & 67.64 & 55.75 & 36.12 \\
  \bottomrule
\end{tabular}
\caption{Feature sets and combinations}
\label{tab:retrieval_feature_combinations}
\end{subtable}

\end{minipage}

\caption{Retrieval performance comparison. (a) Layer types. (b) Connection methods. (c) Feature sets and combinations.}
\label{tab:retrieval_all_combined}
\end{table}

Looking at the individual features, namely S, K, P, and W, the low retrieval scores for the keypoint positions and especially the word embeddings indicate that these features alone do not provide meaningful discriminative information. SIFT descriptors, on the other hand, already show relevant results with a \gls{map} of almost 10\% and a Top-1 result of about 20\% with 283 possible classes during retrieval. The image patch embeddings, however, stand out as the best-performing among the individual features with a mAP of 58.79\%. %It is therefore reasonable to assume that combinations of features without image patch embeddings are unlikely to be relevant. This assumption is confirmed when looking at all the possible combinations of 2 such features.
Furthermore, the results indicate that both word embeddings and the SIFT feature descriptors enhance the performance of the image patches, while the inclusion of the keypoint position leads to a decrease in performance. %This negative impact on the results can be seen in all combinations that include the keypoint positions when compared to the results of the same combination without it, except for KW, which slightly outperforms word embeddings by themselves. With a \gls{map} of just above 2\% that result is, however, negligible for practical applications.
%Among the tested combinations, SKPW does not achieve the highest results but still performs competitively, achieving a \gls{map} of 52.31\% compared to the best overall configuration's (SP) 60.09\%. This difference is considered acceptable to maintain a standardized setup for comparability across experiments. It is therefore kept as the default feature combination for subsequent experiments, as it includes all available feature types.

\subsubsection{Graph Connection Methods - 1Line vs. KNN vs. Full}
\label{subsec:connection_methods}
This section focuses on the results of different connection methods used during the graph-building process, which are introduced in Section \ref{sec:graph_building}.

The results presnted in Table \ref{tab:retrieval_connection_modes} indicate that both 1Line and KNN achieve similar performance, with 1Line achieving slightly higher scores across all metrics. The Full connection method, which introduces a much higher number of edges compared to the other two methods, performs worse by about 15\% and 17\% \gls{map}, respectively. This suggests that an edge density beyond a certain point negatively affects the model's ability to learn meaningful structural relationships.

\subsubsection{Graph Scope - Lines vs. Document}
For the document-level experiments, a graph is constructed for each document by sequentially processing each word in order and adding its corresponding nodes from all lines in a document, effectively creating one continuous line.
The rest of the setup and training process follows the line-level experiments for the first part of this section and is then adapted to handle the lower sample size in the document-level task in the second part. An excerpt of the feature combinations is evaluated, selecting the top-performing combination in each category from previous experiments based on the number of features combined, and including S as the basis for the image patches. Table \ref{tab:line_doc_results} shows the results of this experiment. The Top-5 and Top-10 results of the per-document experiments are always 0, since only 5 documents per author exist, and one of them is used as the current sample.

\begin{table}[t]
\centering

\begin{subtable}[b]{0.48\columnwidth}
\centering

\begin{tabular}{lccc}
\toprule
 & \multicolumn{3}{c}{Retrieval (\%)} \\
\cmidrule{2-4}
 & mAP & Top-1 & Top-3 \\
\midrule
S   & 22.67 & 28.18 & 3.05 \\
P   & 78.98 & 87.79 & 66.50 \\
\textbf{SP}  & \textbf{79.36} & \textbf{88.64} & \textbf{66.36} \\
SPW & 78.05 & 87.08 & 64.37 \\
SKPW& 56.78 & 71.19 & 30.73 \\
\bottomrule
\end{tabular}
\caption{ }
\label{tab:line_doc_results}
\end{subtable}
\hfill
\begin{subtable}[b]{0.48\columnwidth}
\centering

\begin{tabular}{lccc}
\toprule
 & \multicolumn{3}{c}{Retrieval (\%)} \\
\cmidrule{2-4}
 & mAP & Top-1 & Top-3 \\
\midrule
S    & 37.96 & 48.97 & 12.49 \\
P    & 76.70 & 85.45 & 62.67 \\
SP   & 78.33 & 87.44 & 63.73 \\
\textbf{SPW}  & \textbf{79.34} & \textbf{88.15} & \textbf{65.08} \\
SKPW & 63.00 & 75.66 & 39.25 \\
\bottomrule
\end{tabular}
\caption{ }
\label{tab:line_doc_kfold}
\end{subtable}

\caption{Performance metrics on document-level graphs with (a) regular training and (b) adjusted, sequential training on the splits. (Top-5 and Top-10 are zero due to the number of pages per writer)}
\label{tab:doc_level_comparison}
\end{table}

The results show that the higher-performing configurations, namely P, SP, SPW, see an increase of about 20\% \gls{map} on document-level compared to the line-level, while SKPW shows only a small increase of approximately 4\%. The worst performing configuration, S, sees an improvement of over 10\% \gls{map}.
Looking at the Top-1 and Top-3 metrics, per-document performances are generally lower than per-line ones, with the exception of S, which shows a higher Top-1 metric, but falls behind on the Top-3 one, following in line with the other configurations. The differences between the better-performing configurations are much less. Taking P as an example, per-document Top-1 percentage is only about 2\% worse, while the Top-3 percentage difference increases to about 10\%.
For SKPW, which is in terms of \gls{map} over 20 percentage points lower than the best performing one, SP, the Top-1 and Top-3 percentages decrease by \textasciitilde 13\% and 37\%, respectively. This indicates that the presence of K, the only feature in SKPW not included in any of the better-performing configurations, negatively affects the model's ability to rank documents correctly. While this effect is already observed in previous experiments in Section \ref{subsec:feature-combinations}, it is amplified at the document-level.
The inverse trends observed between \gls{map} and Top-k percentages are a consequence of the limited number of samples. This sample limitation, potentially in combination with the current training method, contributes to the per-document results not reaching state-of-the-art performance.

To improve results with a fixed number of 7 samples per author during training, the data for each class is split into all possible 5-2-folds, using 5 samples for training and 2 for validation, resulting in 21 different splits. The model is trained sequentially on each split without resetting, using the updated weights from the previous split for the next iteration. The results of this training approach are shown in Table \ref{tab:line_doc_kfold}.

% \begin{table}
%   \centering
%   \scriptsize
%   \begin{tabular}{lc|ccccc}
%     \toprule
%     & \multicolumn{1}{c|}{Training} & \multicolumn{5}{c}{Retrieval (\%)} \\
%     \cmidrule{2-7}
%     & \#Epochs & mAP & Top-1 & Top-3 & Top-5 & Top-10 \\
%     \midrule
%     S    & 423 & 37.96 & 48.97 & 12.49 & 0.00 & 0.00 \\
%     P    & 500 & 76.70 & 85.45 & 62.67 & 0.00 & 0.00 \\
%     SP   & 484 & 78.33 & 87.44 & 63.73 & 0.00 & 0.00 \\
%     \textbf{SPW}  & \textbf{155} & \textbf{79.34} & \textbf{88.15} & \textbf{65.08} & 0.00 & 0.00 \\
%     SKPW & 259 & 63.00 & 75.66 & 39.25 & 0.00 & 0.00 \\
%     \bottomrule
%   \end{tabular}

%   \caption{Performance metrics of the adjusted training method on document-level graphs.}
%   \label{tab:line_doc_kfold}
% \end{table}
% \begin{table}
%   \centering
%   \scriptsize
%   \begin{tabular}{l|ccccc}
%     \toprule
%     & \multicolumn{5}{c}{Retrieval (\%)} \\
%     \cmidrule{2-6}
%     & mAP & Top-1 & Top-3 & Top-5 & Top-10 \\
%     \midrule
%     S    & 37.96 & 48.97 & 12.49 & 0.00 & 0.00 \\
%     P    & 76.70 & 85.45 & 62.67 & 0.00 & 0.00 \\
%     SP   & 78.33 & 87.44 & 63.73 & 0.00 & 0.00 \\
%     \textbf{SPW}  & \textbf{79.34} & \textbf{88.15} & \textbf{65.08} & 0.00 & 0.00 \\
%     SKPW & 63.00 & 75.66 & 39.25 & 0.00 & 0.00 \\
%     \bottomrule
%   \end{tabular}
%   \caption{Performance metrics of the adjusted training method on document-level graphs.}
%   \label{tab:line_doc_kfold}
% \end{table}

The results demonstrate that the better performing configurations do not improve, but instead exhibit a decrease in \gls{map} and Top-k metrics. In contrast, S and SKPW both improved, with S displaying an increase in \gls{map} of \textasciitilde 15\%. Since the epochs of both S and SKPW do not reach their limit during the standard training method, it suggests that the sequential training avoids local minima that these configurations are more prone to fall into. Overall, this approach does not provide a significant gain for the better-performing configurations.

\subsubsection{Custom Layout and Comparison to State of the Art}
\label{sec:comp_sota}
Since the setup used for the preceding tests allows only a limited variety in terms of \gls{gnn} configurations and primarily focuses on establishing a baseline of what is achievable with basic networks of different kinds, this section explores a custom architecture that build upon the learnings from the previously described results.

This experiment uses the GIN convolutional layer with the 1Line connection method, as both performed best previously. Three feature configurations are tested: P, SP, and SKPW. P and SP are included based on prior results, while SKPW is used to evaluate the effect of combining all features in a performance-focused setup rather than as a baseline.

Architectural and hyperparameter changes are explored, including larger hidden dimensions, dropout, additional GIN layers, and deeper MLPs. Only increasing the hidden dimensions leads to a notable performance improvement; all other changes reduce performance or have little effect.

Initial experiments increase hidden dimensions by roughly 30\% per layer, yielding consistent gains. Larger expansions (512 and 1024 hidden units) introduce higher variability, which is mitigated by reducing the learning rate from 0.001 to 0.0001, as confirmed by five runs per feature configuration. With this adjustment, performance improves across all feature sets except one SKPW run (\gls{map} 46.15\%), while most results remain around 64\%.
As shown in Table \ref{tab:custom_retrieval_summary}, all configurations outperform the baseline setups from Section \ref{subsec:feature-combinations} by 3–6 percentage points in \gls{map}. Using only image patch embeddings achieves the best performance (mean \gls{map} 65.14\%) with the lowest variability. Adding SIFT features reduces performance, consistent with baseline results indicating that SIFT does not reliably improve outcomes. The combination of all features performs worst and shows the highest variability, though its best run slightly exceeds SP. Overall, increasing hidden dimensions combined with a lower learning rate improves performance while maintaining robustness.

% \begin{table}
%   \centering
%   \resizebox{\columnwidth}{!}{
%   \begin{tabular}{l@{\hspace{50pt}}r}
%     \toprule
%     \multicolumn{2}{c}{\textbf{Training Configuration}} \\
%     \midrule
%     Learning Rate      & 0.0001 \\
%     Batch Size         & 32 \\
%     Epochs             & 250 \\
%     Patience           & 15 \\
%     \midrule
%     \multicolumn{2}{c}{\textbf{Model Architecture}} \\
%     \midrule
%     Hidden Dimension   & 512 $\rightarrow$ 1024 \\
%     Number of Layers   & 2 \\
%     Layer Type         & GIN \\
%     Features           & P, SP, SKPW \\
%     Connection Mode    & 1Line \\
%     \midrule
%     \multicolumn{2}{c}{\textbf{Feature Dimensions}} \\
%     \midrule
%     SIFT Descriptor (S)           & 128 \\
%     SIFT Keypoint (K)       & 2 \\
%     Image Patch Embedding (P)          & 64 \\
%     Word Embedding (W)  & 101 \\
%     \bottomrule
%   \end{tabular}
%   }

%   Hidden Dimension same as input dimension
% Number of Layers 2
% Layer Type GCN
% Features SKPW
% Connection Mode 1line
% Scope line-level
%   \caption{Hyperparameters and architecture information of the custom layout. Feature dimensions are shown as a reference to calculate layer dimensions.}
%   \label{tab:custom_layout_params}
% \end{table}

\begin{table}
\centering
\begin{tabular}{lcccccc}
\toprule
Setup & Epochs & mAP  & Top-1 & Top-3  & Top-5  & Top-10 \\
\midrule
SKPW & $120\pm15$ & $60.71\pm8.2$ & $92.22\pm2.4$ & $80.93\pm5.2$ & $71.50\pm6.9$ & $53.03\pm8.8$ \\
SP   & $136\pm37$ & $63.01\pm1.3$ & $93.34\pm0.7$ & $82.94\pm1.5$ & $74.14\pm2.0$ & $56.18\pm2.4$ \\
P    & $106\pm18$ & $65.14\pm1.1$ & $94.57\pm0.6$ & $85.19\pm1.1$ & $76.91\pm1.4$ & $59.81\pm1.7$ \\
\bottomrule
\end{tabular}
\caption{Retrieval performance (mean ± standard deviation, in \%) for SKPW, SP, and P over five runs. Best \glspl{map} achieved: SKPW 64.96\%, SP 64.62\%, P 66.58\%. Epoch values are rounded to the nearest integer.}
\label{tab:custom_retrieval_summary}
\end{table}

% The results in Table \ref{tab:custom_retrieval_summary} show that all tested feature combinations perform notably better than the established baseline configurations introduced in Section \ref{subsec:feature-combinations}, reaching 3 to 6 percentage points higher \gls{map}. The setup using only image patch embeddings as input performs best, achieving a mean \gls{map} of 65.14\% across five runs and demonstrating the lowest variability. The combination of image patch embeddings with SIFT descriptors shows a decline in performance compared to the image-patch-only setup, which contrasts with the established baseline experiments. However, as shown in those baseline tests, the inclusion of SIFT features does not consistently improve performance, such as the configurations PW (59.58\% \gls{map}) and SPW (57.96\% \gls{map}), suggesting that SIFT features are not a reliable way to improve results, while remaining relevant for evaluation. The combination of all features performs overall worst and exhibits the highest variability, though its best individual run outperforms SP marginally. These results confirm that the expansion of hidden layer dimensions in tandem with a reduced learning rate yields an overall improvement and provides a balance between performance and robustness.

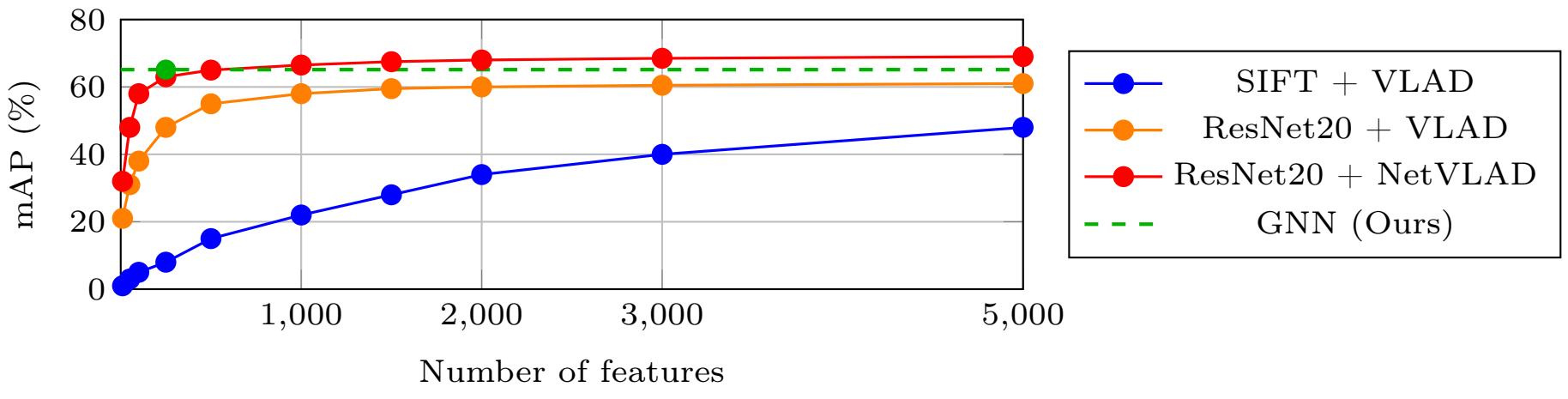
\begin{figure}
    \centering
    \begin{tikzpicture}
    \begin{axis}[
        xlabel={Number of features},
        ylabel={mAP (\%)},
        xmin=0, xmax=5000,
        ymin=0, ymax=80,
        xtick={1000,2000,3000,5000},
        ytick={0,20,40,60,80},
        grid=major,
        width=0.7\columnwidth,    % shrink the plot width
        height=0.30\columnwidth,  % shrink the plot height
        tick label style={font=\scriptsize},   % smaller tick labels
        xlabel style={font=\scriptsize},     % smaller x label
        ylabel style={font=\scriptsize},     % smaller y label
        legend style={font=\scriptsize},     % keep legend readable
        legend style={at={(1.05, 0.5)}, anchor=west, legend columns=1},
        legend columns = 1,
    ]
    
    % Blue Line (Bottom)
    \addplot [semithick, color=blue, mark=*]
    table {%
        10 1.0
        50 3.0
        100 5.0
        250 8.0
        500 15.0
        1000 22.0
        1500 28.0
        2000 34.0
        3000 40.0
        5000 48.0
    };
    \addlegendentry{SIFT + VLAD}
    
    % Orange Line (Middle)
    \addplot [semithick, color=orange, mark=*]
    table {%
        10 21.0
        50 31.0
        100 38.0
        250 48.0
        500 55.0
        1000 58.0
        1500 59.5
        2000 60.0
        3000 60.5
        5000 61.0
    };
    \addlegendentry{ResNet20 + VLAD}
    
    % Red Line (Top)
    \addplot [semithick, color=red, mark=*]
    table {%
        10 32.0
        50 48.0
        100 58.0
        250 63.0
        500 65.0
        1000 66.5
        1500 67.5
        2000 68.0
        3000 68.5
        5000 69.0
    };
    \addlegendentry{ResNet20 + NetVLAD}
    
    % \addplot [only marks, color=teal, mark=*]
    % table {%
    % 250 66.58
    % };
    % \addplot [dashed, teal] coordinates {(0,66.58) (5000,66.58)};
    % \addlegendentry{GNN (top)}

    % \addplot [only marks, color=green!70!black, mark=*]
    % table {%
    % 250 65.14
    % };
    \addplot [dashed, thick, green!70!black] coordinates {(0,65.14) (5000,65.14)};
    \addlegendentry{GNN (Ours)}

    % % GNN (top) dot at x = 250
    % \addplot [only marks, color=teal, mark=*]
    % coordinates {(250,66.58)};
    
    % GNN (mean) dot at x = 250
    \addplot [only marks, color=green!70!black, mark=*]
    coordinates {(250,65.14)};
        
    \end{axis}
    \end{tikzpicture}
    \caption{Comparison of the proposed \gls{gnn}-based approach, using an average of \textasciitilde250 nodes per line of text, with state-of-the-art methods on the CVL dataset \cite{peer2025towards}. The dotted lines serve as an additional indicator to support comparison with state-of-the-art methods at different numbers of features.}
    \label{fig:sota_comparison}
\end{figure}

\paragraph{Comparison with state of the art} Figure \ref{fig:sota_comparison} illustrates a comparison of the proposed approach against existing state-of-the-art methods presented by Peer et al. \cite{peer2025towards} on the CVL dataset. The original results include SIFT+VLAD, ResNet20+VLAD, and ResNet20+NetVLAD configurations, while the teal line indicates the performance of the best \gls{gnn}-based result using only image patch embeddings as input feature, and the green line shows its mean value.
As shown in the figure, the proposed method performs best for a low number of features. At around 250 feature points per line, it outperforms all state-of-the-art approaches, with ResNet20+NetVLAD surpassing the mean of the proposed approach at around 500 features, and surpassing its best run at around 1000 feature points. This demonstrates that the \gls{gnn}-based aggregation method is a viable approach for \gls{wr}, particularly when using a lower number of feature points.

\subsection{Identification}
\label{sec:indentification}
%This section introduces the experiments conducted to evaluate \gls{wi}.
For these experiments, the \gls{cnn} used to extract image patch embeddings is trained on the IAM dataset. This prevents data leakage, as \gls{wi} requires the same authors to appear in both training and testing, necessitating a split of the original CVL training set. Training the \gls{cnn} on the IAM dataset, therefore, removes the need to retrain it whenever new train–test splits for the \gls{gnn} are created.
The \gls{wi} experiments use the same standard configuration as the retrieval experiments shown in Table \ref{tab:training_config}. All settings remain identical unless a particular parameter is intentionally varied to evaluate its effect, such as the choice of layer type.

\subsubsection{Layer Types - GCN, GAT, GATv2, SAGE, GIN}
\label{subsec:layer_types_id}
Similar to the tests during retrieval this experiment investigates the performance of different layer types. Table \ref{tab:test_acc_layers} shows the results.

% \begin{table}
%   \centering
%   \scriptsize
%   \begin{tabular}{lc|c|c|c|c}
%     \toprule
%     & \multicolumn{3}{c|}{Training Metrics} & \multicolumn{2}{c}{Identification Metrics} \\
%     \cmidrule{2-6}
%     & \# Epochs & Train Loss & Val Loss & Test Loss & Test Accuracy (\%) \\
%     \midrule
%     GCN   & 99  & 0.0061 & 0.0283 & 0.2067 & 97.4 \\
%     GAT   & 107 & 0.0195 & 0.0555 & 0.2341 & 97.0 \\
%     GATv2 & 116 & 0.0125 & 0.0507 & 0.2060 & 97.8 \\
%     SAGE  & 64  & 0.0192 & 0.0826 & 0.1637 & 98.1 \\
%     \textbf{GIN}   & \textbf{43}  & \textbf{0.0109} & \textbf{0.0177} & \textbf{0.0575} & \textbf{99.6} \\
%     \bottomrule
%   \end{tabular}
%   \caption{Training and identification performance metrics of different layer types}
%   \label{tab:training_identification_layer_results}
% \end{table}
% \begin{table}
%   \centering
%   \scriptsize
%   \begin{tabular}{l|c|c|c|c}
%     \toprule
%     & \multicolumn{2}{c|}{Training Metrics} & \multicolumn{2}{c}{Identification Metrics} \\
%     \cmidrule{2-5}
%     & Train Loss & Val Loss & Test Loss & Test Accuracy (\%) \\
%     \midrule
%     GCN   & 0.0061 & 0.0283 & 0.2067 & 97.4 \\
%     GAT   & 0.0195 & 0.0555 & 0.2341 & 97.0 \\
%     GATv2 & 0.0125 & 0.0507 & 0.2060 & 97.8 \\
%     SAGE  & 0.0192 & 0.0826 & 0.1637 & 98.1 \\
%     \textbf{GIN}   & \textbf{0.0109} & \textbf{0.0177} & \textbf{0.0575} & \textbf{99.6} \\
%     \bottomrule
%   \end{tabular}
%   \caption{Training and identification performance metrics of different layer types.}
%   \label{tab:training_identification_layer_results}
% \end{table}

\begin{table}[t]
\centering

\begin{subtable}[b]{0.32\linewidth}
\centering
\begin{tabular}{lc}
  \toprule
  & Test Accuracy (\%) \\
  \midrule
  GCN   & 97.4 \\
  GAT   & 97.0 \\
  GATv2 & 97.8 \\
  SAGE  & 98.1 \\
  \textbf{GIN} & \textbf{99.6} \\
  \bottomrule
\end{tabular}
\caption{Layer types}
\label{tab:test_acc_layers}
\end{subtable}
\hfill
\begin{subtable}[b]{0.32\linewidth}
\centering
\begin{tabular}{lc}
  \toprule
  & Test Accuracy (\%) \\
  \midrule
  S    & 79.18 \\
  \textbf{P}   & \textbf{99.26} \\
  \textbf{SP}  & \textbf{99.26} \\
  SPW  & 97.40 \\
  SKPW & 97.03 \\
  \bottomrule
\end{tabular}
\caption{Feature combinations}
\label{tab:test_acc_features}
\end{subtable}
\hfill
\begin{subtable}[b]{0.32\linewidth}
\centering
\begin{tabular}{lc}
  \toprule
  & Test Accuracy (\%) \\
  \midrule
  1Line & 96.28 \\
  \textbf{KNN}  & \textbf{96.65} \\
  \textbf{Full} & \textbf{96.65} \\
  \bottomrule
\end{tabular}
\caption{Connection methods}
\label{tab:test_acc_connections}
\end{subtable}

\caption{Test accuracy comparison. (a) Identification accuracy of different GNN layer types. (b) Identification accuracy of different feature sets and their combinations. (c) Identification accuracy of different connection methods.}
\label{tab:test_accuracy_comparison}
\end{table}

All layer types have a test accuracy of over 97\%, and exactly like the retrieval results, GCN is ahead of GAT and behind all others. Furthermore, GIN is again, just like during retrieval, the best performing layer type with a test accuracy of 99.6\%.
    
\subsubsection{Feature Combinations - S/K/P/W}
\label{subsec:feature-combinations_id}
This experiment evaluates different feature combinations during \gls{wi}. As shown in the corresponding \gls{wr} experiments in Section \ref{subsec:feature-combinations}, the keypoint position feature K decreases performance and is therefore excluded from the subset of feature combinations used here. SKPW is tested as it serves as the standard configuration for all experiments. Table \ref{tab:test_acc_features} shows the results.

Both P and SP achieve the highest test accuracy of 99.26\%. The SIFT descriptors alone achieve the lowest accuracy, but have no negative impact on performance when combined with the image patch embeddings. On their own SIFT descriptors reach almost 80\% test accuracy, which is notable since their retrieval performance, shown in Section \ref{subsec:feature-combinations}, only reaches about 10\% \gls{map}. The results of SPW and SKPW indicate that the inclusion of K has no positive impact, as the observed difference is marginal, with under 1 percentage point, which can be attributed to the individual trained models and not the method itself.

\subsubsection{Graph Connection Methods - 1Line vs. KNN vs. Full}
\label{subsec:connection_methods_id}
%fk%The connection mode performances during the \gls{wr} experiments are similar for 1Line and KNN and lower for Full, as the additional information hinders the training process. Similar results are expected for this identification experiment, which are shown in Table \ref{tab:training_identification_connection_results}.

% \begin{table}
%   \centering
%   \scriptsize
%   \begin{tabular}{lc|c|c|c|c}
%     \toprule
%     & \multicolumn{3}{c|}{Training Metrics} & \multicolumn{2}{c}{Identification Metrics} \\
%     \cmidrule{2-6}
%     & \# Epochs & Train Loss & Val Loss & Test Loss & Test Accuracy (\%) \\
%     \midrule
%     1Line & 55 & 0.0344 & 0.1037 & 0.2898 & 96.28 \\
%     \textbf{KNN}   & \textbf{48} & \textbf{0.0415} & \textbf{0.0486} & \textbf{0.1038} & \textbf{96.65} \\
%     \textbf{Full}  & \textbf{76} & \textbf{0.0068} & \textbf{0.2114} & \textbf{0.3659} & \textbf{96.65} \\
%     \bottomrule
%   \end{tabular}
%   \caption{Training and identification performance metrics of the 3 different connection methods}
%   \label{tab:training_identification_connection_results}
% \end{table}

%fk%Contrary to the assumptions, t
The results of all connection methods shown in Table \ref{tab:test_acc_connections} are nearly identical, with accuracies of almost 97\% and only negligible differences. Since identification experiments that include image patch embeddings achieve results close to 100\% without parameter optimization or combining the best settings of the individual experiments, it can be inferred that, based on the data used, the task of \gls{wi} is easier than retrieval and that the choice between IAM and CVL trained CNNs has little impact. The results demonstrate that the use of \glspl{gnn} for \gls{wi} is effective. 

%% file: sections/conclusion.tex
\section{Conclusion}

This work investigates \glspl{gnn} as learned aggregation mechanisms for offline, text-independent \gls{wi} and \gls{wr}, representing handwritten documents and lines as graphs whose nodes encode local handwriting features and whose edges model spatial relations. Experiments on the CVL and IAM datasets demonstrate that \gls{gnn}-based aggregation achieves performance competitive with state-of-the-art pooling-based methods, particularly for \gls{wr} under conditions with limited numbers of feature points, while maintaining near state-of-the-art accuracy for \gls{wi}. Among the evaluated features, \gls{cnn}-based image patch embeddings yield the strongest and most stable results. Sparse graph construction strategies that preserve spatial structure, such as 1Line and KNN, consistently outperform fully connected graphs for \gls{wr}, whereas \gls{wi} proves largely insensitive to the chosen connectivity. %Across different \gls{gnn} architectures, most layer types perform similarly, with GIN providing notable improvements in both retrieval and identification tasks. 
Overall, the results highlight the effectiveness of graph-based representations and \gls{gnn}-based aggregation for capturing fine-grained handwriting characteristics, especially in low-information regimes, and establish a flexible framework for future extensions in feature design and graph construction.

Future work should investigate improved feature extraction and graph construction strategies. Dimensionality reduction methods such as PCA can be applied to high-dimensional descriptors, and the number of keypoints may be increased to better capture writer-specific characteristics; additional structural features and weighted edges are also worth-investigating. Furthermore, alternative graph construction schemes, including sparsified and multi-level graphs that integrate word-, line-, and document-level information, might be considerable to better preserve spatial relations and reduce structural bias.

\section{Acknowledgements}

This research was funded in whole or in part by the Austrian Science Fund (FWF) 10.55776/PAT4919325. Marco Peer was partly supported by the Swiss National Science Foundation (SNSF) under project grant no. 10002280, “AI-Supported Human-Computer Interaction with Large Volumes of Historical Documents".